%% file: main.tex
\documentclass[10pt,twocolumn,letterpaper]{article}

\usepackage{iccv}              %

\input{preamble}
\definecolor{iccvblue}{rgb}{0.21,0.49,0.74}
\usepackage[pagebackref,breaklinks,colorlinks,allcolors=iccvblue]{hyperref}

\usepackage{graphicx}
\usepackage{multirow}
\usepackage{colortbl}
\usepackage{epsfig} %

\input{mathlib.tex}

\graphicspath{{./figs/}{./figs/teaser/}{./figs/sketch/}{./figs/eval/}}

\newcommand\blfootnote[1]{%
  \begingroup
  \renewcommand\thefootnote{}\footnote{#1}%
  \addtocounter{footnote}{-1}%
  \endgroup
}

\def\paperID{14602} %
\def\confName{ICCV}
\def\confYear{2025}

\title{Relative Illumination Fields:\\ Learning Medium and Light Independent Underwater Scenes}

\author{
Mengkun She$^{1*}$
\and 
Felix Seegr\"aber$^{1*}$
\and
David Nakath$^{1}$
\and
Patricia Sch\"ontag$^{2}$%
\and 
Kevin K\"oser$^{1}$
\and
\\
\begin{tabular}{cc}
    $^{1}$Institute of Computer Science & $^{2}$GEOMAR\\
    Kiel University, Germany & Helmholtz Centre for Ocean Research Kiel, Germany\\
\end{tabular}~\\
}

\begin{document}
\twocolumn[{%
\renewcommand\twocolumn[1][]{#1}%
\maketitle
\vspace{-3em}
\def\svgwidth{0.99\textwidth}
\begin{center}
    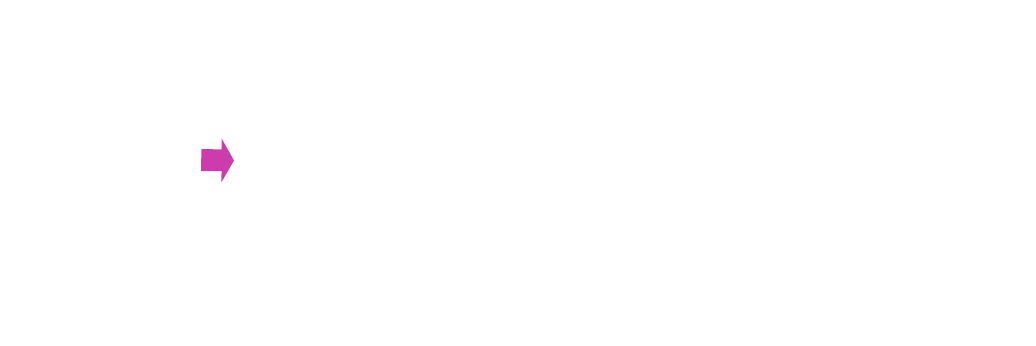
\end{center}
\captionof{figure}{Our proposed approach reconstructs a neural radiance field of an underwater scene captured by a camera system with unknown, co-moving light sources, enabling novel view synthesis (a). Additionally, it can recover the clean scene representation where neither medium nor light cone effects are present (b) and is capable of disentangling the light cone in the scattering medium (c) and the surface illumination (d).}
\label{fig:teaser}
}]
\maketitle
{\blfootnote{*Authors contributed equally to this work.}}

\input{sec/0_abstract}    
\input{sec/1_intro}

\input{sec/2_related_work}
\input{sec/3_method}

\input{sec/4_experiments}
\input{sec/5_conclusion}
{
    \small
    \bibliographystyle{ieeenat_fullname}
    \bibliography{main}
}
\setcounter{section}{0}

\end{document}

%% file: mathlib.tex
\renewcommand{\d}[1]{\mbox{\boldmath$#1$}}

\newcommand{\q}[1]{{{\bf #1}}}

\newcommand{\mq}[1]{{\q #1}}

\newcommand{\mC}{\mathchoice {\setbox0=\hbox{$\displaystyle\rm
C$}\hbox{\hbox to0pt{\kern0.4\wd0\vrule height0.9\ht0\hss}\box0}}
{\setbox0=\hbox{$\textstyle\rm C$}\hbox{\hbox
to0pt{\kern0.4\wd0\vrule height0.9\ht0\hss}\box0}}
{\setbox0=\hbox{$\scriptstyle\rm C$}\hbox{\hbox
to0pt{\kern0.4\wd0\vrule height0.9\ht0\hss}\box0}}
{\setbox0=\hbox{$\scriptscriptstyle\rm C$}\hbox{\hbox
to0pt{\kern0.4\wd0\vrule height0.9\ht0\hss}\box0}}}

\newcommand{\mG}{\mathchoice {\setbox0=\hbox{$\displaystyle\rm
G$}\hbox{\hbox to0pt{\kern0.4\wd0\vrule height0.9\ht0\hss}\box0}}
{\setbox0=\hbox{$\textstyle\rm G$}\hbox{\hbox
to0pt{\kern0.4\wd0\vrule height0.9\ht0\hss}\box0}}
{\setbox0=\hbox{$\scriptstyle\rm G$}\hbox{\hbox
to0pt{\kern0.4\wd0\vrule height0.9\ht0\hss}\box0}}
{\setbox0=\hbox{$\scriptscriptstyle\rm G$}\hbox{\hbox
to0pt{\kern0.4\wd0\vrule height0.9\ht0\hss}\box0}}}

\newcommand{\mJ}{\mathchoice {\setbox0=\hbox{$\displaystyle\rm
J$}\hbox{\hbox to0pt{\kern0.4\wd0\vrule height0.9\ht0\hss}\box0}}
{\setbox0=\hbox{$\textstyle\rm J$}\hbox{\hbox
to0pt{\kern0.4\wd0\vrule height0.9\ht0\hss}\box0}}
{\setbox0=\hbox{$\scriptstyle\rm J$}\hbox{\hbox
to0pt{\kern0.4\wd0\vrule height0.9\ht0\hss}\box0}}
{\setbox0=\hbox{$\scriptscriptstyle\rm J$}\hbox{\hbox
to0pt{\kern0.4\wd0\vrule height0.9\ht0\hss}\box0}}}

\newcommand{\mO}{\mathchoice {\setbox0=\hbox{$\displaystyle\rm
O$}\hbox{\hbox to0pt{\kern0.4\wd0\vrule height0.9\ht0\hss}\box0}}
{\setbox0=\hbox{$\textstyle\rm O$}\hbox{\hbox
to0pt{\kern0.4\wd0\vrule height0.9\ht0\hss}\box0}}
{\setbox0=\hbox{$\scriptstyle\rm O$}\hbox{\hbox
to0pt{\kern0.4\wd0\vrule height0.9\ht0\hss}\box0}}
{\setbox0=\hbox{$\scriptscriptstyle\rm O$}\hbox{\hbox
to0pt{\kern0.4\wd0\vrule height0.9\ht0\hss}\box0}}}

\newcommand{\mQ}{\mathchoice {\setbox0=\hbox{$\displaystyle\rm
Q$}\hbox{\raise 0.15\ht0\hbox to0pt{\kern0.4\wd0\vrule
height0.8\ht0\hss}\box0}}{\setbox0=\hbox{$\textstyle\rm Q$}\hbox{\raise
0.15\ht0\hbox to0pt{\kern0.4\wd0\vrule height0.8\ht0\hss}\box0}}
{\setbox0=\hbox{$\scriptstyle\rm Q$}\hbox{\raise 0.15\ht0\hbox
to0pt{\kern0.4\wd0\vrule height0.7\ht0\hss}\box0}}{\setbox0=
\hbox{$\scriptscriptstyle\rm Q$}\hbox{\raise 0.15\ht0\hbox
to0pt{\kern0.4\wd0\vrule height0.7\ht0\hss}\box0}}}

\newcommand{\mS}{\mathchoice
{\setbox0=\hbox{$\displaystyle     \rm S$}\hbox{\raise0.5\ht0\hbox
to0pt{\kern0.35\wd0\vrule height0.45\ht0\hss}\hbox
to0pt{\kern0.55\wd0\vrule height0.5\ht0\hss}\box0}}
{\setbox0=\hbox{$\textstyle        \rm S$}\hbox{\raise0.5\ht0\hbox
to0pt{\kern0.35\wd0\vrule height0.45\ht0\hss}\hbox
to0pt{\kern0.55\wd0\vrule height0.5\ht0\hss}\box0}}
{\setbox0=\hbox{$\scriptstyle      \rm S$}\hbox{\raise0.5\ht0\hbox
to0pt{\kern0.35\wd0\vrule height0.45\ht0\hss}\raise0.05\ht0\hbox
to0pt{\kern0.5\wd0\vrule height0.45\ht0\hss}\box0}}
{\setbox0=\hbox{$\scriptscriptstyle\rm S$}\hbox{\raise0.5\ht0\hbox
to0pt{\kern0.4\wd0\vrule height0.45\ht0\hss}\raise0.05\ht0\hbox
to0pt{\kern0.55\wd0\vrule height0.45\ht0\hss}\box0}}}
\newcommand{\mT}{\mathchoice {\setbox0=\hbox{$\displaystyle\rm
T$}\hbox{\hbox to0pt{\kern0.3\wd0\vrule height0.9\ht0\hss}\box0}}
{\setbox0=\hbox{$\textstyle\rm T$}\hbox{\hbox
to0pt{\kern0.3\wd0\vrule height0.9\ht0\hss}\box0}}
{\setbox0=\hbox{$\scriptstyle\rm T$}\hbox{\hbox
to0pt{\kern0.3\wd0\vrule height0.9\ht0\hss}\box0}}
{\setbox0=\hbox{$\scriptscriptstyle\rm T$}\hbox{\hbox
to0pt{\kern0.3\wd0\vrule height0.9\ht0\hss}\box0}}}
\newcommand{\mU}{\mathchoice {\setbox0=\hbox{$\displaystyle\rm
U$}\hbox{\hbox to0pt{\kern0.4\wd0\vrule height0.9\ht0\hss}\box0}}
{\setbox0=\hbox{$\textstyle\rm U$}\hbox{\hbox
to0pt{\kern0.4\wd0\vrule height0.9\ht0\hss}\box0}}
{\setbox0=\hbox{$\scriptstyle\rm U$}\hbox{\hbox
to0pt{\kern0.4\wd0\vrule height0.9\ht0\hss}\box0}}
{\setbox0=\hbox{$\scriptscriptstyle\rm U$}\hbox{\hbox
to0pt{\kern0.4\wd0\vrule height0.9\ht0\hss}\box0}}}

%% file: figs/teaser/teaser_v2.pdf_tex
\begingroup%
  \makeatletter%
  \providecommand\color[2][]{%
    \errmessage{(Inkscape) Color is used for the text in Inkscape, but the package 'color.sty' is not loaded}%
    \renewcommand\color[2][]{}%
  }%
  \providecommand\transparent[1]{%
    \errmessage{(Inkscape) Transparency is used (non-zero) for the text in Inkscape, but the package 'transparent.sty' is not loaded}%
    \renewcommand\transparent[1]{}%
  }%
  \providecommand\rotatebox[2]{#2}%
  \newcommand*\fsize{\dimexpr\f@size pt\relax}%
  \newcommand*\lineheight[1]{\fontsize{\fsize}{#1\fsize}\selectfont}%
  \ifx\svgwidth\undefined%
    \setlength{\unitlength}{496.85998535bp}%
    \ifx\svgscale\undefined%
      \relax%
    \else%
      \setlength{\unitlength}{\unitlength * \real{\svgscale}}%
    \fi%
  \else%
    \setlength{\unitlength}{\svgwidth}%
  \fi%
  \global\let\svgwidth\undefined%
  \global\let\svgscale\undefined%
  \makeatother%
  \begin{picture}(1,0.33409815)%
    \lineheight{1}%
    \setlength\tabcolsep{0pt}%
    \put(0,0){\includegraphics[width=\unitlength,page=1]{teaser_v2.pdf}}%
    \put(-0.00172805,0.00674902){\color[rgb]{0,0,0}\makebox(0,0)[lt]{\lineheight{1.25}\smash{\begin{tabular}[t]{l}Multi-view input images\end{tabular}}}}%
    \put(0.42917876,0.00674902){\color[rgb]{0,0,0}\makebox(0,0)[lt]{\lineheight{1.25}\smash{\begin{tabular}[t]{l}(b) Color restored\end{tabular}}}}%
    \put(0.24664059,0.00675095){\color[rgb]{0,0,0}\makebox(0,0)[lt]{\lineheight{1.25}\smash{\begin{tabular}[t]{l}(a) Re-rendering\end{tabular}}}}%
    \put(0.60665725,0.00674902){\color[rgb]{0,0,0}\makebox(0,0)[lt]{\lineheight{1.25}\smash{\begin{tabular}[t]{l}(c) Medium backscatter\end{tabular}}}}%
    \put(0.80580315,0.00674902){\color[rgb]{0,0,0}\makebox(0,0)[lt]{\lineheight{1.25}\smash{\begin{tabular}[t]{l}(d) Surface illumination\end{tabular}}}}%
    \put(0,0){\includegraphics[width=\unitlength,page=2]{teaser_v2.pdf}}%
  \end{picture}%
\endgroup%

%% file: sec/0_abstract.tex
\begin{abstract}
We address the challenge of constructing a consistent and photorealistic Neural Radiance Field in inhomogeneously illuminated, scattering environments with unknown, co-moving light sources. While most existing works on underwater scene representation focus on a static homogeneous illumination, limited attention has been paid to  scenarios such as when a robot explores water deeper than a few tens of meters, where sunlight becomes insufficient.
To address this, we propose a novel illumination field locally attached to the camera, enabling the capture of uneven lighting effects within the viewing frustum. We combine this with a volumetric medium representation to an overall method that effectively handles interaction between dynamic illumination field and static scattering medium. 
Evaluation results demonstrate the effectiveness and flexibility of our approach. 
\end{abstract}

%% file: sec/1_intro.tex
\section{Introduction}
\label{sec:intro}

Neural Radiance Fields (NeRFs)~\cite{mildenhall2020nerf, barron2022mip}, along with their subsequent variants~\cite{tancik2023nerfstudio, kerbl3Dgaussians}, have demonstrated remarkable capabilities in generating high-fidelity, photorealistic renderings of scenes reconstructed from 2D images. 
Although NeRF’s formulation is volumetric, it assumes the scene exists in a clear air environment with highly opaque objects.
Underwater environments, however, present unique challenges due to wavelength-dependent attenuation and scattering. 
The space between objects is no longer empty but filled with water, which absorbs and scatters light, behaving almost like an additional object.
This interaction complicates the disentanglement of true objects from the medium, making the current NeRF formulation less effective.
Recent advancements~\cite{levy2023seathru, tang2024neural, li2024watersplatting} in underwater scene representation have adapted NeRF-based methods for these situations. 
Nevertheless, most approaches focus on homogeneous scattering medium illuminated by a distant global light source, such as sunlight. 
These methods primarily rely on the revised underwater imaging formation model~\cite{akkaynak2018revised,akkaynak2019sea}, which is largely based on the well-known fog model~\cite{nayar1999vision} for shallow water scenarios, where sunlight produces nearly isotropic veiling light, closely resembling in-air fog conditions.

The dark depths of the ocean, with little or no natural light, cover more than half of Earth's surface but remains largely uncharted and unexplored. Marine robots almost always carry their own light sources, since even in shallower waters artificial light is needed for bad weather, turbid waters or night time. 
These inhomogeneous and dynamic illumination sources interact with the static scattering medium, creating strong inhomogeneous backscatter light cones~\cite{koser2020challenges, Song2022}.
\begin{figure}
    \centering
    \includegraphics[width=0.95\columnwidth]{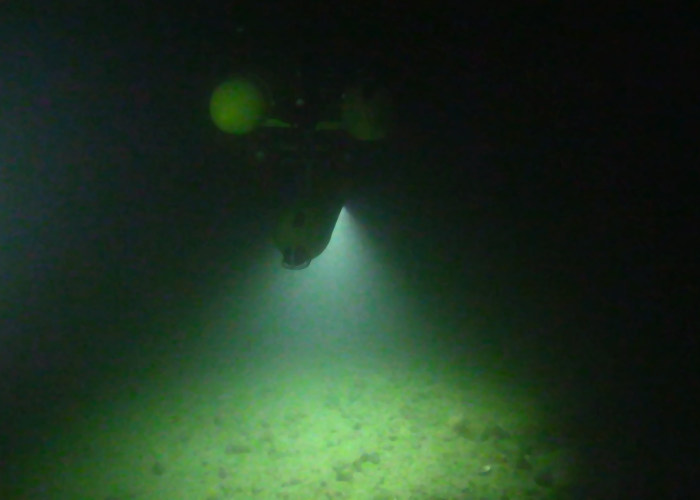}
    \caption{Autonomous underwater vehicle in $100m$ depth in coastal waters. Note the illumination pattern at the seafloor as well as the scattering (the light cone) in the water.
    }
    \label{fig:cams_example}
\end{figure}
We find that the state-of-the-art underwater NeRF approaches cannot easily handle images taken in such scenarios, and only a handful of previous works partially address these challenges. 
Relightable NeRFs~\cite{zeng2023nrhints,zhang2021nerfactor} try to factorize illumination from a scene representation for relighting purposes, and these methods are typically demonstrated in synthetic environments with a single object centered in the image with a clean background, or in an artificially controlled real-world setups.
DarkGS~\cite{zhang2024darkgs} tackles co-moving light sources in dark, in-air environments, constructing an illumination-consistent Gaussian representation of the scene.
Classical methods addressing artificial lights in a  scattering medium~\cite{bryson2016true,  nakath2021situ, song2024advanced} typically require prior radiometric calibration, which limits their practical usability, or rely on strong assumptions about the lights.

The closest work to ours in the literature is probably~\cite{zhang2023beyond}, which assumes a single, ideal point light source at the camera center. Since practical underwater systems try to position light sources as far from the camera as possible~\cite{mcglamery1980computer}, our work drops~\cite{zhang2023beyond}'s restriction. Additionally, nowadays' robots are often equipped with many adjustable and spatially distributed LED spotlights that make accurate pose measurement and explicit parameterization of all the lights challenging. Therefore, we propose a \emph{non-parametric illumination field} representation that is locally \emph{attached to the camera}.
By integrating the underwater imaging formulation into the rendering process, we achieve a clean scene representation, invariant to both illumination and medium effects, as shown in Fig. \ref{fig:teaser}. 
The entire pipeline is optimized jointly without requiring any calibration of the light. 
Our key contributions are as follows

\begin{itemize}
    \item A local camera viewing frustum-based MLP to model the inhomogeneous illumination field, which dynamically co-moves with the camera.
    \item A volumetric radiance field representation capable of managing the interaction of the co-moving light field with the absorbing and scattering medium.
    \item Extensive evaluations using realistically generated synthetic images (physically-based volumetric path-tracer) and real-world datasets to validate the proposed approach.    
\end{itemize}

%% file: sec/2_related_work.tex
\section{Related Work}
\label{sec:related_work}

\paragraph{Neural Radiance Fields}

NeRF~\cite{mildenhall2020nerf} has emerged as a powerful tool for photorealistic view synthesis by fitting an implicit scene representation to calibrated multi-view images. Subsequent variations, such as Instant-NGP~\cite{muller2022instant} and Mip-NeRF~\cite{barron2021mip}, have improved both rendering quality and training speed. 
RawNeRF~\cite{mildenhall2022nerf} enables High Dynamic Range (HDR) view synthesis through training on dark, RAW-format input images.
Capturing neither over nor under exposed images in low-light environments with artificial lighting requires an exceptionally high dynamic range. As a result, we follow the techniques proposed in RawNeRF and train our model using RAW images.

To make NeRF more robust for real-world captures, NeRF-Wild~\cite{martin2021nerf} introduces image-based appearance embeddings and optimizes the latent space to account for images taken at different times of day under varying illumination conditions. However, our goal goes beyond building a consistent representation—we aim to restore a clean representation by removing the effects of both lighting and the medium.
NeRFactor~\cite{zhang2021nerfactor} trains a neural reflectance field alongside BRDF, light visibility fields, and surface normals, allowing for the factorization of a clean scene representation under a single unknown lighting condition. 
S$^3$-NeRF~\cite{yang2022s3nerf} exploits shading and shadow information to estimate the neural reflectance field, assuming the light position is known.
DarkGS~\cite{zhang2024darkgs} is one of the few works to address co-moving light source scenarios.  Although it is based on 3D Gaussian Splatting (3DGS), its core approach involves learning the Radiant Intensity Distribution (RID) of the light and the fall-off curve using MLPs. A radiometric calibration step is required to obtain this information, along with the light pose.

\paragraph{Underwater Color Restoration}
Early works from Preisendorfer~\cite{preisendorfer1976hydrologic} have looked into the low level physics of light transport at an infinitesimally small volume of water.
Jaffe and McGlamery~\cite{jaffe1990computer,mcglamery1980computer} suggests to decompose the underwater imaging formation process into direct signals, forward-scatter and back-scatter components.
Effectively, a fraction of light is \textit{absorbed}, while another is \textit{scattered} to another direction.
Which direction the light is scattered is governed by Volume Scattering Function (VSF), which is a directional distribution function that varies according to the local water composition~\cite{petzold1972volume}.

To restore the true color of the scene from underwater photos, the image formation process needs to be inverted. 
But the full physical model is computationally expensive and difficult to invert.
For a more uniform illumination scenario, Akkaynak and Treibitz~\cite{akkaynak2018revised,akkaynak2019sea} introduce a revised underwater imaging formation model, along with the SeaThru approach which estimates the model parameters using RGB-D data.
Their approach is also later integrated into NeRF~\cite{levy2023seathru,tang2024neural} and 3DGS~\cite{li2024watersplatting}.
Interestingly, Tang et al.~\cite{tang2024neural} introduce an MLP to compensate globally inhomogeneous light distribution within the water column.

For scenarios involving artificial lighting, Bryson et al.~\cite{bryson2016true} employ a physical model that utilizes a pre-calibrated Gaussian light source to restore colors accurately. 
Nakath et al.~\cite{nakath2021situ} and Song et al.~\cite{song2024advanced} both present an in-situ calibration-based approach for color restoration.
The first employs a Monte-Carlo volumetric path tracer~\cite{nimier2019mitsuba} to first optimize the light parameters and the medium-related parameters, then to restore the object's texture.
The latter pre-calibrates a 3D grid structure, storing multiplicative and additive factors for each grid cell that determine the pixel color underwater.
The work most similar to ours is~\cite{zhang2023beyond}, which learns a color-restored neural reflectance field under a single point light source located at the camera center. Assuming the light source is co-located at the camera center allows simple application of inverse-square and Lambert’s cosine law. Our approach differs by imposing no restrictions on the light sources, including their number, position, orientation or intensity profile. Moreover, our method is calibration-free, learning the distribution of light intensity in space from the data.

%% file: sec/3_method.tex
\section{Method}\label{sec:method}

Our goal is to construct a NeRF of the underlying scene with the effects of lights and medium removed, using a collection of calibrated multi-view images (e.g. obtained from underwater SfM\cite{she24rsfm}). 
NeRF utilizes a multilayer perceptron (MLP), denoted as $\mathcal F_\Theta$ , to learn a continuous volumetric radiance field within a bounded 3D volume. 
The MLP takes as input a positional-encoded vector $\phi_{\mathrm{Hash}}(\d x)$ and a directional-encoded vector $\phi_{\mathrm{SH}}(\d d)$, and outputs the density $\sigma$ at position $\d x = (x, y, z)$ and the corresponding color $\d c = (r, g, b)$ observed along the viewing direction $\d d = (d_x, d_y, d_z)$.
Here, $\phi_{\mathrm{Hash}} (\cdot)$ denotes the encoding function for positions and $\phi_{\mathrm{SH}} (\cdot)$ for the unit-norm viewing directions.
In this work, $\phi_{\mathrm{Hash}}(\cdot)$ is chosen as the multi-resolution hash encoding proposed by Instant-NGP~\cite{muller2022instant} and $\phi_{\mathrm{SH}}(\cdot)$ is simply the Spherical-Harmonics encoding.

Next, to render the pixel color $\d C(\d r(t))$ of a particular ray $\d r(t) = \d o + t \d d$ originating from the projection center $\d o$, NeRF uses the volumetric ray-marching algorithm~\cite{max1995optical} in the approximated discretized form:
\begin{gather}\label{eq:nerf_render}
    \d C(\d r(t)) = \sum_i^N T_i (1 - \exp(-\sigma_i \delta_i)) \d c_i\\
    \mathrm{where}\;\; T_i = \exp(-\sum_j^{i-1} \sigma_j \delta_j), \nonumber
\end{gather}
$t$ is the distance to a sample and $\delta_i = t_{i+1} - t_i$ is the distance between two samples points. 
$T_i$ denotes the accumulated transmittance from the beginning of the ray to the current sample position.
The model is then trained by minimizing the image reconstruction loss for all the rays:
\begin{equation}\label{eq:loss_basic}
    \mathcal{L} = \sum_{\d r \in \mathcal{R}} \Vert \hat{C}(\d r(t)) - C(\d r(t)) \Vert^2
\end{equation}
Images captured in dark, low light environments often show a very high dynamic range. The fall-off characteristics of artificial illumination, further amplified by medium attenuation, necessitates varying camera shutter speeds to allow the sensor to work in its ideal range. 
This further increases the dynamic range over the dataset. 
Since varying shutter speeds effectively change the sensitivity of the camera sensor we either require knowledge about the exposure time of each view or an overall consistent shutter speed. To deal with high dynamic range and varying exposure times we adopt the strategies outlined in RawNeRF~\cite{mildenhall2022nerf}. The loss function in Eq. \ref{eq:loss_basic} becomes:
\begin{equation}
    \mathcal{L} = \sum_{\d r \in \mathcal{R}} \Vert \frac{\hat{C}(\d r(t)) - C(\d r(t))}{\mathrm{sg}(\hat{C}(\d r(t))) + \epsilon} \Vert^2
\end{equation}
where $\mathrm{sg}(\cdot)$ stands for stop-gradient, and $\epsilon = 1\cdot10^{-3}$. 

\begin{figure*}[!ht]
    \centering
    \footnotesize
    \def\svgwidth{0.98\textwidth}
    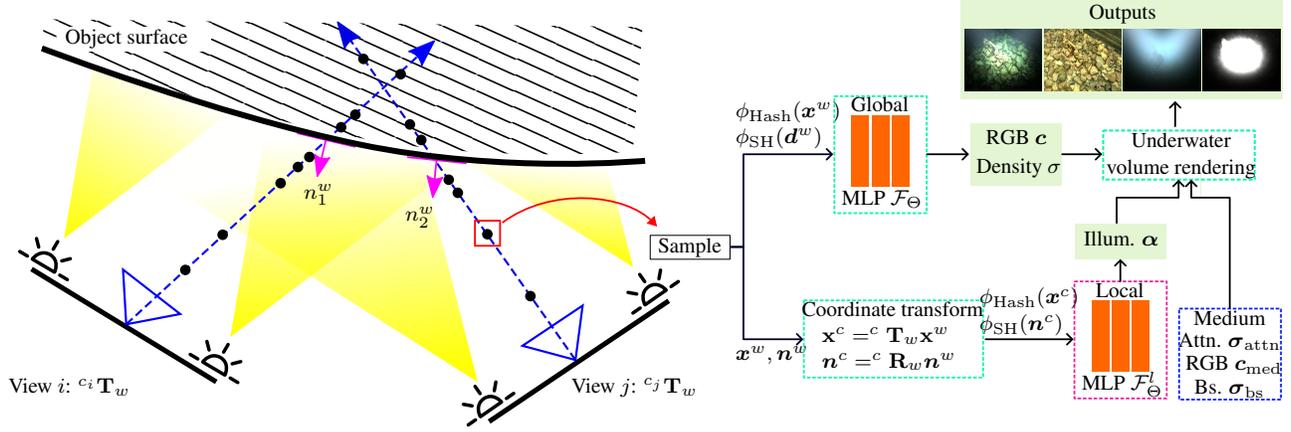
    \caption{An illustration of the problem setup and the architecture of our proposed approach. The global NeRF MLP $\mathcal{F}_{\Theta}$ learns both density and color at each ray sample in the world coordinate frame. Simultaneously, each ray sample is transformed into the local camera coordinate frame, with surface normals derived from the predicted density field, allowing the local illumination field MLP $\mathcal{F}^l_{\Theta}$ to estimate the light intensity factor $\d \alpha$ that the sample point receives. Finally, medium-related parameters, that is the attenuation coefficient $\d \sigma_{\mathrm{attn}}$, medium color $\d c_{\mathrm{med}}$, and backscatter $\d \sigma_{\mathrm{bs}}$ are jointly estimated and integrated together into a underwater volume rendering formulation.}
    \label{fig:sketch}
\end{figure*}

\subsection{Local Illumination Field Representation}

In dark environments a camera is often accompanied by a mobile light to provide the necessary illumination. The resulting extreme brightness variation of objects in the scene is challenging for reconstruction algorithms, because it breaks the photometric consistency between different views. A common approach in this scenario is to explicitly consider the light sources to compensate this effect.

Previous studies have demonstrated that lights can be modeled in various forms, such as Gaussian lights, projectors with texture patterns, or lights with neural network-predicted RID and fall-off curves ~\cite{nakath2021situ, zhang2024darkgs, zhang2023beyond}. The advantage of these models is that the pixel color can be accurately determined using ray-tracing, which is physically accurate and effectively captures shadow effects.
However, a key limitation of these methods is the requirement for known light numbers, models and poses. 
Jointly estimating all light parameters along with the scene within a differentiable ray-tracing framework has been shown to be challenging~\cite{nakath2021situ}. Consequently, a calibration step is often necessary. Additionally, ray-tracing in a volumetric medium requires accurate modeling of the Volume Scattering Function.

In the co-moving light source scenario, we observe that the cumulative illumination a surface point receives within the camera’s viewing frustum is more critical than knowing all the characteristics of each individual light source. This is because the three dimensional illumination pattern remains constant in relation to the camera (see Fig. \ref{fig:teaser}). Observing this pattern on the same scene over several views allows to draw conclusions about the distribution of light intensity in space and its relation to the object reflectance.
Therefore, inspired by NeRF, we propose to fit a network to the input data to “memorize” the light intensity distribution in the local camera frame.
To this end, we employ a simple MLP $\mathcal{F}^l_\Theta$ that maps a positional-encoded vector $\phi_\mathrm{Hash}(\d x^c)$, but in the local camera coordinate frame to a light intensity factor $\d \alpha$, representing the illumination received at that point.
Here, the superscripts $^l$ and $^c$ denote the light field and the coordinate frame respectively.
 
However, we understand that the light intensity received by a surface point should also depend on the local surface normal $\d n^c$ and the incoming light direction $\d d_l$, as well as the visibility of the light source. 
The reflected light at this point further depends on its material properties and spectral reflectance.
Consequently, the radiance $\d c_i$ emitted by a sample point in Eq.~\ref{eq:nerf_render} should be replaced by:
\begin{equation}
    \d c_i^\prime = \d \alpha_i(\d x^c, \d n^c, \d d_l) \cdot f_r \cdot \lambda_{\mathrm{vis}} \cdot \d c_i
\end{equation}
where $\lambda_{\mathrm{vis}}$ represents a binary function indicating whether this sample is visible to the light, $f_r$ denotes the Bidirectional Reflectance Distribution Function (BRDF), and $\d c_i$ corresponds to the reflectance (albedo), which is the same as the original NeRF representation.

Although \cite{zhang2021nerfactor} has shown that it is feasible to learn the BRDF and a visibility mask within the NeRF framework, we choose to omit it to reduce network complexity, given that most non-artificial materials can be reasonably assumed to behave Lambertian underwater\cite{bryson2016true}. 
Moreover, we also drop the light visibility term, as the lights are commonly close to the camera and shadows are cast behind objects, outside of view. Neglecting shadows admittedly limits our ability to model them when present, so we acknowledge this as a limitation and suggest it be addressed in future work.
Therefore, the above equation can be simplified to:
\begin{equation}
    \d c^\prime_i = \d \alpha_i(\d x^c, \d n^c, \d d_l) \cdot \d c_i
\end{equation}
In our representation, we are not aware of the number, the location or the intensity profile of the light sources. What matters is that the relative direction of the surface normal, within the local camera frame, remains consistent with respect to the light source. We therefore drop the $\d d_l$ term and provide the local surface normal $\d n^c$ as an input to the network to let it determine the light intensity a given sample point should receive. Hence, the final local illumination field representation is:
\begin{equation}\label{eq:alpha}
    \d \alpha = \mathcal{F}^l_\Theta(\phi_\mathrm{Hash}(\d x^c), \phi_\mathrm{SH}(\d n^c))
\end{equation}
Representing camera pose as $^{c} \mq T_w = [^c\mq R_w \;|\; ^c \d t_w]$\footnote{The transformation matrix $^b\mq T_a$ transforms a point in the $a$ coordinate frame to the $b$ coordinate frame. Elements of projective space are written with upright serifs, e.g. $\q x \in \mathbb{P}^3$, and the Euclidean elements are written in \textit{italics} such as $\d x \in \mathbb{R}^3$.},  point $\d x^c$ and normal $\d n^c$ can be obtained via transformation:
\begin{equation}
    \q x^c = ^c\mq T_w \cdot \q x^w \;\;\mathrm{and}\;\; \d n^c = ^c\mq R_w \cdot \d n^w
\end{equation}
the surface normal in the world coordinate frame $\d n^w$ can be determined by calculating the gradient of the predicted density field with respect to the positions~\cite{verbin2022ref}.
We introduce $\d x^w$ to indicate the 3D position of a point in the world coordinate frame.
But if the point is a sample along the camera viewing ray, then $\d x^w = \d r(t) = \d o + t \d d$.

\subsection{Medium Representation}

To model the effects of the medium between camera and objects we employ a volume rendering formulation akin to the one utilized in NeRF (Eq.~\ref{eq:nerf_render}), but adding a medium component into it:
\begin{gather}
    \d C(\d r(t)) = \nonumber \\ 
    \sum_i^N T_i (1 - e^{-(\sigma^\mathrm{obj}_i + \sigma^{\mathrm{med}}_i) \delta_i}) \frac{\sigma^{\mathrm{obj}}_i \d c^\mathrm{obj}_i + \sigma_i^{\mathrm{med}} \d c^\mathrm{med}_i }{\sigma_i^{\mathrm{obj}} + \sigma_i^{\mathrm{med}}} \\
    \mathrm{where}\;\; T_i = \exp(-\sum_j^{i-1} (\sigma^{\mathrm{obj}}_j + \sigma^{\mathrm{med}}_j ) \delta_j) \nonumber
\end{gather}
and the original $(\sigma_i, \d c_i)$ are replaced by $(\sigma_i^\mathrm{obj}, \d c^{\mathrm{obj}}_i)$ to distinguish between the object and the medium.
This formulation was already used in~\cite{levy2023seathru,tang2024neural} where they show that it aligns with previous image formation models for fog or in water.
As suggested in~\cite{levy2023seathru}, the rendered color of a pixel can be decomposed into the object color (direct signal) and the medium color (backscatter component) assuming that object and medium density do not intersect, i.e. $\sigma^{\mathrm{obj}}_i \gg \sigma^{\mathrm{med}}_i$ if the sample is at the surface of the object and $\sigma^{\mathrm{obj}}_i \ll \sigma^{\mathrm{med}}_i$ if the sample is in the medium ~\cite{akkaynak2018revised,akkaynak2019sea}:
\begin{equation}\label{eq:medium_render}
    \d C(\d r(t)) = \sum_i^N \d C_i^{\mathrm{obj}}(\d r(t)) + \d C_i^{\mathrm{med}}(\d r(t))
\end{equation}
with
\begin{align*}
      \d C_i^{\mathrm{obj}}(\d r(t)) &= T^{\rm obj}_i \cdot T^{\rm attn}_i \cdot \big(1-\exp({-\sigma^{\rm obj}_i\delta_i})\big) \cdot \mathbf{c}^{\rm obj}_i \\
      \d C_i^{\mathrm{med}}(\d r(t)) &= T^{\rm obj}_i\cdot T^{\rm bs}_i \cdot \big(1-\exp({-\d \sigma^{\rm bs}\delta_i})\big) \cdot  \mathbf{c}^{\rm med} 
\end{align*}
\begin{equation*}      
    \mathrm{where}\;\; T^{\rm type}_i = \exp\bigg(-\sum_{j=0}^{i-1}\sigma^{\rm type}_j\delta_j\bigg)
\end{equation*}
and $\rm type$ is one of $\rm obj, attn$ or  $\rm bs$ for object, attenuation and backscatter respectively. 
For attenuation and backscatter this simplifies to $\exp\left(-{\boldsymbol{\sigma}} s_i\right)$ with $s_i = \sum_{j=0}^{i-1}\delta_j$  because $\d \sigma^{\rm attn}$ and $\d \sigma^{\rm bs}$ are considered constant over the whole ray.
Due to our formulation we can drop the view dependency used in \cite{levy2023seathru} and consider the medium to be constant over the whole medium. We also do not use a medium MLP but rather implement $\d \sigma^{\rm attn}, \d \sigma^{\rm bs}$ and $\d c^{\rm med}$ as single optimizable parameters.

\subsection{Final Model}

Our framework is illustrated in Fig.~\ref{fig:sketch}. Combining \cref{eq:medium_render} with the illumination field Eq.~\ref{eq:alpha} is straight forward. The factor $\d \alpha_i$ acts on both the medium and object color samples which simplifies to:
\begin{equation}\label{eq:final_model}
    \d C(\d r(t)) = \sum_i^N \d \alpha_i \bigg( C_i^{\mathrm{obj}}(\d r(t)) + \d C_i^{\mathrm{med}}(\d r(t)) \bigg)
\end{equation}
Note that this model is also applicable to in-air scenes with co-moving light sources by simply omitting $\d C^{\mathrm{med}}(\d r(t))$. 
In a medium the light is not only attenuated on its path from the object to the camera but also prior to that, on its way from its source to the object. The above formulation only considers the distance from camera to object since its the one that corresponds to the sample depth. For a single point light source located at the camera center the observed attenuation of the light could be modeled by simply doubling the actual medium attenuation. This is because the distance $t^c$ and $t^l$ of camera and light to $\d x^c$ are equal and the attenuation over the whole path of the light becomes $\exp(-\d \sigma^{\mathrm{attn}} t^l) \cdot \exp(-\d \sigma^{\mathrm{attn}} t^c) = \exp(-2 \cdot \d \sigma^{\mathrm{attn}}  t^c)$. For any other light configuration the relation between $t^c$ and $t^l$ depends on the relative poses of the co-moving light sources to the camera and the location of $\d x^c$. We therefore estimate an $\alpha$ per color channel to allow the illumination field to represent the cumulative attenuated light received at each point $\d x^c$ for arbitrary light configurations.
\subsection{Implementation Details}

\begin{figure*} [t]
    \centering
    \def\svgwidth{0.9\textwidth}
    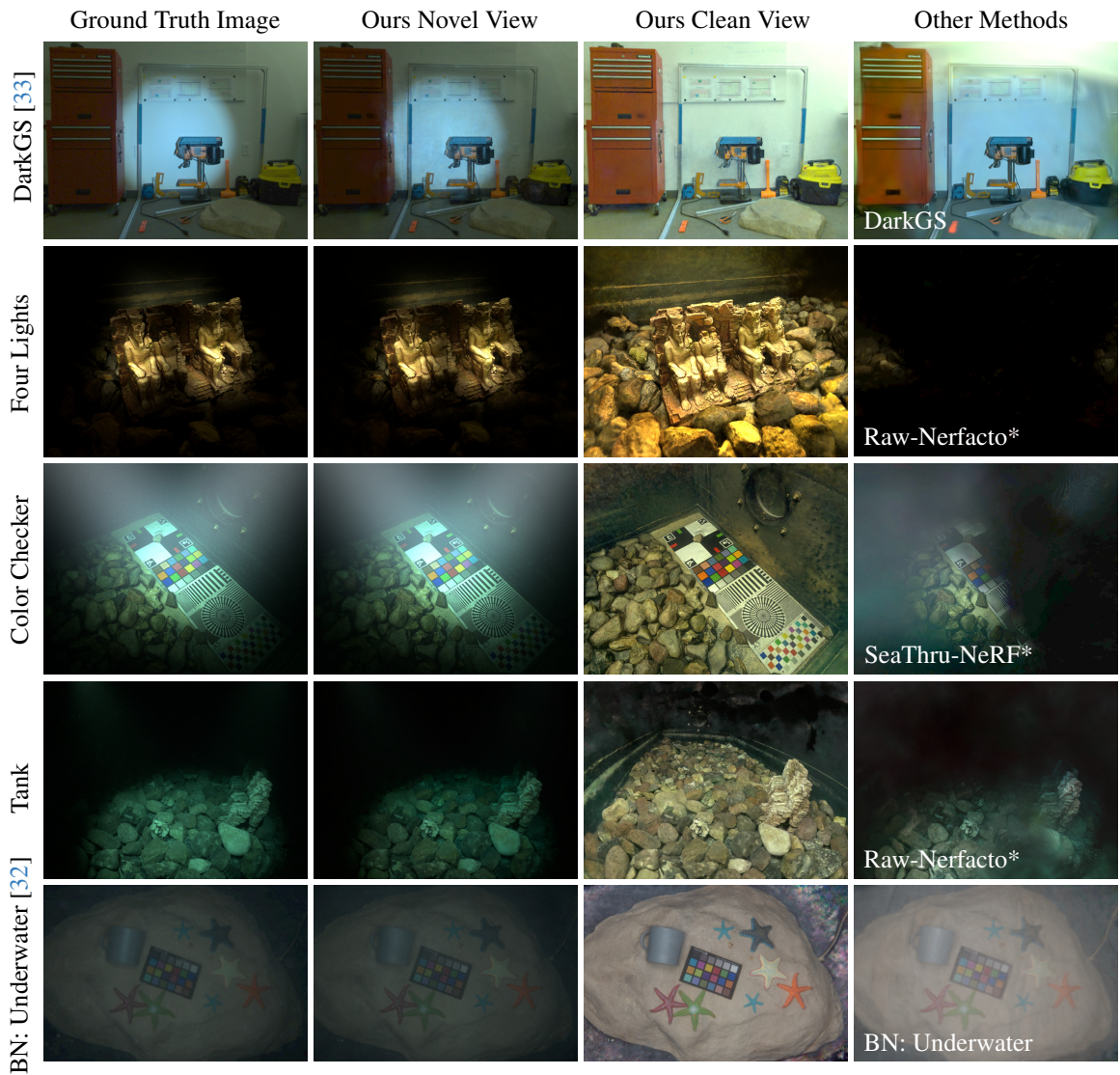
    \caption{Experiments on different datasets. Top two rows are in air, bottom three in water. Sets contain one, four, two, two and one co-moving light from top to bottom. The \textit{Four Lights} and \textit{Color Checker} dataset are rendered, the \textit{Tank} dataset is captured by us, the DarkGS \cite{zhang2024darkgs} and Beyond Nerf: Underwater \cite{zhang2023beyond} datasets are freely available. Since we are tackling a largely unsolved problem, there are virtually no competitor methods to compare against. We found it still instructive to present results obtained by the methods marked with an asterisk that do not explicitly model co-moving light but are still related to our scenario in some aspects.}
    \label{fig:eval_method}
\end{figure*}

Our method is implemented as an extension to the Nerfstudio framework \cite{tancik2023nerfstudio}.
We use the Nerfacto method from Nerfstudio as our baseline NeRF, which includes features like proposal network sampling, multi-resolution hash encoding and scene contraction from~\cite{barron2022mip,muller2022instant}.
However, we disable the NeRF-Wild~\cite{martin2021nerf} appearance embedding, which we find to interfere with our light representation and the camera pose optimization. 
As mentioned before we also employ the strategies from RawNeRF to deal with HDR imagery: Input data is normalized to a common exposure level and loss is calculated on RAW data masked with the corresponding bayer pattern for each channel. We also follow in using an exponential activation function for the color output of the scene MLP but use a sigmoid activation for the illumination field. We find that our method performs better when introducing a dataset depending scaling factor on $\d \alpha$. Our assumption is that this allows to deal with the different dynamic ranges on the individual sets but requires further investigation.
Lastly, we also apply an exponential activation function to $\d c^{\rm{med}}$ and a softplus one to $\d \sigma^{\rm{attn}}, \d \sigma^{\rm{bs}}$ even though they are single parameters, ensuring they remain within the range $(0, \infty)$ .

%% file: figs/sketch/sketch_v2.pdf_tex
\begingroup%
  \makeatletter%
  \providecommand\color[2][]{%
    \errmessage{(Inkscape) Color is used for the text in Inkscape, but the package 'color.sty' is not loaded}%
    \renewcommand\color[2][]{}%
  }%
  \providecommand\transparent[1]{%
    \errmessage{(Inkscape) Transparency is used (non-zero) for the text in Inkscape, but the package 'transparent.sty' is not loaded}%
    \renewcommand\transparent[1]{}%
  }%
  \providecommand\rotatebox[2]{#2}%
  \newcommand*\fsize{\dimexpr\f@size pt\relax}%
  \newcommand*\lineheight[1]{\fontsize{\fsize}{#1\fsize}\selectfont}%
  \ifx\svgwidth\undefined%
    \setlength{\unitlength}{510.23622047bp}%
    \ifx\svgscale\undefined%
      \relax%
    \else%
      \setlength{\unitlength}{\unitlength * \real{\svgscale}}%
    \fi%
  \else%
    \setlength{\unitlength}{\svgwidth}%
  \fi%
  \global\let\svgwidth\undefined%
  \global\let\svgscale\undefined%
  \makeatother%
  \begin{picture}(1,0.33888889)%
    \lineheight{1}%
    \setlength\tabcolsep{0pt}%
    \put(0,0){\includegraphics[width=\unitlength,page=1]{sketch_v2.pdf}}%
    \put(0.23958757,0.18309916){\color[rgb]{0,0,0}\makebox(0,0)[lt]{\lineheight{1.25}\smash{\begin{tabular}[t]{l}$n^w_1$\end{tabular}}}}%
    \put(0.31849927,0.16548836){\color[rgb]{0,0,0}\makebox(0,0)[lt]{\lineheight{1.25}\smash{\begin{tabular}[t]{l}$n^w_2$\end{tabular}}}}%
    \put(0.0102021,0.0316302){\color[rgb]{0,0,0}\makebox(0,0)[lt]{\lineheight{1.25}\smash{\begin{tabular}[t]{l}View $i$: $^{c_i}\mq T_w$\end{tabular}}}}%
    \put(0.44748404,0.0316302){\color[rgb]{0,0,0}\makebox(0,0)[lt]{\lineheight{1.25}\smash{\begin{tabular}[t]{l}View $j$: $^{c_j}\mq T_w$\end{tabular}}}}%
    \put(0,0){\includegraphics[width=\unitlength,page=2]{sketch_v2.pdf}}%
    \put(0.53935926,0.1409957){\color[rgb]{0,0,0}\makebox(0,0)[t]{\lineheight{1.25}\smash{\begin{tabular}[t]{c}Sample \end{tabular}}}}%
    \put(0,0){\includegraphics[width=\unitlength,page=3]{sketch_v2.pdf}}%
    \put(0.57514882,0.24629242){\color[rgb]{0,0,0}\makebox(0,0)[lt]{\lineheight{1.25}\smash{\begin{tabular}[t]{l}$\phi_{\mathrm{Hash}}(\d x^w)$\\$\phi_{\mathrm{SH}}(\d d^w)$\end{tabular}}}}%
    \put(0,0){\includegraphics[width=\unitlength,page=4]{sketch_v2.pdf}}%
    \put(0.68710742,0.25055583){\color[rgb]{0,0,0}\makebox(0,0)[t]{\lineheight{1.25}\smash{\begin{tabular}[t]{c}Global\end{tabular}}}}%
    \put(0,0){\includegraphics[width=\unitlength,page=5]{sketch_v2.pdf}}%
    \put(0.65758277,0.17815017){\color[rgb]{0,0,0}\makebox(0,0)[lt]{\lineheight{1.25}\smash{\begin{tabular}[t]{l}MLP $\mathcal{F}_\Theta$\end{tabular}}}}%
    \put(0,0){\includegraphics[width=\unitlength,page=6]{sketch_v2.pdf}}%
    \put(0.57514682,0.05732837){\color[rgb]{0,0,0}\makebox(0,0)[lt]{\lineheight{1.25}\smash{\begin{tabular}[t]{l}$\d x^w, \d n^w$\end{tabular}}}}%
    \put(0,0){\includegraphics[width=\unitlength,page=7]{sketch_v2.pdf}}%
    \put(0.69703183,0.09120801){\color[rgb]{0,0,0}\makebox(0,0)[t]{\lineheight{1.25}\smash{\begin{tabular}[t]{c}Coordinate transform\end{tabular}}}}%
    \put(0.64208115,0.07171906){\color[rgb]{0,0,0}\makebox(0,0)[lt]{\lineheight{1.25}\smash{\begin{tabular}[t]{l}$\q x^c = ^c\mq T_w \q x^w$\end{tabular}}}}%
    \put(0.64208115,0.05144878){\color[rgb]{0,0,0}\makebox(0,0)[lt]{\lineheight{1.25}\smash{\begin{tabular}[t]{l}$\d n^c = ^c\mq R_w \d n^w$\end{tabular}}}}%
    \put(0,0){\includegraphics[width=\unitlength,page=8]{sketch_v2.pdf}}%
    \put(0.87358466,0.10654182){\color[rgb]{0,0,0}\makebox(0,0)[t]{\lineheight{1.25}\smash{\begin{tabular}[t]{c}Local\end{tabular}}}}%
    \put(0.95884221,0.08468658){\color[rgb]{0,0,0}\makebox(0,0)[t]{\lineheight{1.25}\smash{\begin{tabular}[t]{c}Medium\end{tabular}}}}%
    \put(0.95907008,0.06765815){\color[rgb]{0,0,0}\makebox(0,0)[t]{\lineheight{1.25}\smash{\begin{tabular}[t]{c}Attn. $\d \sigma_{\mathrm{attn}}$\end{tabular}}}}%
    \put(0.96231842,0.04945513){\color[rgb]{0,0,0}\makebox(0,0)[t]{\lineheight{1.25}\smash{\begin{tabular}[t]{c}RGB $\d c_{\mathrm{med}}$\end{tabular}}}}%
    \put(0.95880093,0.03105426){\color[rgb]{0,0,0}\makebox(0,0)[t]{\lineheight{1.25}\smash{\begin{tabular}[t]{c}Bs. $\d \sigma_{\mathrm{bs}}$\end{tabular}}}}%
    \put(0,0){\includegraphics[width=\unitlength,page=9]{sketch_v2.pdf}}%
    \put(0.84405975,0.03413616){\color[rgb]{0,0,0}\makebox(0,0)[lt]{\lineheight{1.25}\smash{\begin{tabular}[t]{l}MLP $\mathcal{F}^l_\Theta$\end{tabular}}}}%
    \put(0,0){\includegraphics[width=\unitlength,page=10]{sketch_v2.pdf}}%
    \put(0.7931054,0.22559214){\color[rgb]{0,0,0}\makebox(0,0)[t]{\lineheight{1.25}\smash{\begin{tabular}[t]{c}RGB $\d c$\end{tabular}}}}%
    \put(0.795091,0.20268897){\color[rgb]{0,0,0}\makebox(0,0)[t]{\lineheight{1.25}\smash{\begin{tabular}[t]{c}Density $\sigma$\end{tabular}}}}%
    \put(0,0){\includegraphics[width=\unitlength,page=11]{sketch_v2.pdf}}%
    \put(0.87476611,0.14644504){\color[rgb]{0,0,0}\makebox(0,0)[t]{\lineheight{1.25}\smash{\begin{tabular}[t]{c}Illum. $\d \alpha$\end{tabular}}}}%
    \put(0,0){\includegraphics[width=\unitlength,page=12]{sketch_v2.pdf}}%
    \put(0.92109892,0.22259103){\color[rgb]{0,0,0}\makebox(0,0)[t]{\lineheight{1.25}\smash{\begin{tabular}[t]{c}Underwater\\volume rendering\end{tabular}}}}%
    \put(0,0){\includegraphics[width=\unitlength,page=13]{sketch_v2.pdf}}%
    \put(0.87584059,0.32252791){\color[rgb]{0,0,0}\makebox(0,0)[t]{\lineheight{1.25}\smash{\begin{tabular}[t]{c}Outputs\end{tabular}}}}%
    \put(0,0){\includegraphics[width=\unitlength,page=14]{sketch_v2.pdf}}%
    \put(0.7646041,0.10126534){\color[rgb]{0,0,0}\makebox(0,0)[lt]{\lineheight{1.25}\smash{\begin{tabular}[t]{l}$\phi_{\mathrm{Hash}}(\d x^c)$\\$\phi_{\mathrm{SH}}(\d n^c)$\end{tabular}}}}%
    \put(0,0){\includegraphics[width=\unitlength,page=15]{sketch_v2.pdf}}%
    \put(0.05429912,0.3032426){\color[rgb]{0,0,0}\makebox(0,0)[lt]{\lineheight{1.25}\smash{\begin{tabular}[t]{l}Object surface\end{tabular}}}}%
    \put(0,0){\includegraphics[width=\unitlength,page=16]{sketch_v2.pdf}}%
  \end{picture}%
\endgroup%

%% file: figs/eval/eval_figure.pdf_tex
\begingroup%
  \makeatletter%
  \providecommand\color[2][]{%
    \errmessage{(Inkscape) Color is used for the text in Inkscape, but the package 'color.sty' is not loaded}%
    \renewcommand\color[2][]{}%
  }%
  \providecommand\transparent[1]{%
    \errmessage{(Inkscape) Transparency is used (non-zero) for the text in Inkscape, but the package 'transparent.sty' is not loaded}%
    \renewcommand\transparent[1]{}%
  }%
  \providecommand\rotatebox[2]{#2}%
  \newcommand*\fsize{\dimexpr\f@size pt\relax}%
  \newcommand*\lineheight[1]{\fontsize{\fsize}{#1\fsize}\selectfont}%
  \ifx\svgwidth\undefined%
    \setlength{\unitlength}{496.85998535bp}%
    \ifx\svgscale\undefined%
      \relax%
    \else%
      \setlength{\unitlength}{\unitlength * \real{\svgscale}}%
    \fi%
  \else%
    \setlength{\unitlength}{\svgwidth}%
  \fi%
  \global\let\svgwidth\undefined%
  \global\let\svgscale\undefined%
  \makeatother%
  \begin{picture}(1,0.94594053)%
    \lineheight{1}%
    \setlength\tabcolsep{0pt}%
    \put(0.06409928,0.91828834){\color[rgb]{0,0,0}\makebox(0,0)[lt]{\lineheight{1.25}\smash{\begin{tabular}[t]{l}Ground Truth Image\end{tabular}}}}%
    \put(0.31957832,0.91765045){\color[rgb]{0,0,0}\makebox(0,0)[lt]{\lineheight{1.25}\smash{\begin{tabular}[t]{l}Ours Novel View\end{tabular}}}}%
    \put(0.02675306,-0.00005912){\color[rgb]{0,0,0}\rotatebox{90}{\makebox(0,0)[lt]{\lineheight{0}\smash{\begin{tabular}[t]{l}BN: Underwater~\cite{zhang2023beyond}\end{tabular}}}}}%
    \put(0.02977341,0.76449743){\color[rgb]{0,0,0}\rotatebox{90}{\makebox(0,0)[lt]{\lineheight{1.25}\smash{\begin{tabular}[t]{l}DarkGS~\cite{zhang2024darkgs}\end{tabular}}}}}%
    \put(0.0272721,0.58026884){\color[rgb]{0,0,0}\rotatebox{90}{\makebox(0,0)[lt]{\lineheight{1.25}\smash{\begin{tabular}[t]{l}Four Lights\end{tabular}}}}}%
    \put(0.0272721,0.37734861){\color[rgb]{0,0,0}\rotatebox{90}{\makebox(0,0)[lt]{\lineheight{1.25}\smash{\begin{tabular}[t]{l}Color Checker\end{tabular}}}}}%
    \put(0.0272721,0.22860204){\color[rgb]{0,0,0}\rotatebox{90}{\makebox(0,0)[lt]{\lineheight{1.25}\smash{\begin{tabular}[t]{l}Tank\end{tabular}}}}}%
    \put(0.55999041,0.91765045){\color[rgb]{0,0,0}\makebox(0,0)[lt]{\lineheight{1.25}\smash{\begin{tabular}[t]{l}Ours Clean View\end{tabular}}}}%
    \put(0,0){\includegraphics[width=\unitlength,page=1]{eval_figure.pdf}}%
    \put(0.80452277,0.91765045){\color[rgb]{0,0,0}\makebox(0,0)[lt]{\lineheight{1.25}\smash{\begin{tabular}[t]{l}Other Methods\end{tabular}}}}%
    \put(0,0){\includegraphics[width=\unitlength,page=2]{eval_figure.pdf}}%
    \put(0.76007035,0.74213039){\color[rgb]{1,1,1}\makebox(0,0)[lt]{\lineheight{1.25}\smash{\begin{tabular}[t]{l}DarkGS\end{tabular}}}}%
    \put(0.76007035,0.02048709){\color[rgb]{1,1,1}\makebox(0,0)[lt]{\lineheight{0}\smash{\begin{tabular}[t]{l}BN: Underwater\end{tabular}}}}%
    \put(0.76007035,0.18129598){\color[rgb]{1,1,1}\makebox(0,0)[lt]{\lineheight{0}\smash{\begin{tabular}[t]{l}Raw-Nerfacto*\end{tabular}}}}%
    \put(0.76073857,0.36092633){\color[rgb]{1,1,1}\makebox(0,0)[lt]{\lineheight{0}\smash{\begin{tabular}[t]{l}SeaThru-NeRF*\end{tabular}}}}%
    \put(0.76007035,0.55185993){\color[rgb]{1,1,1}\makebox(0,0)[lt]{\lineheight{0}\smash{\begin{tabular}[t]{l}Raw-Nerfacto*\end{tabular}}}}%
  \end{picture}%
\endgroup%

%% file: sec/4_experiments.tex
\section{Experiments}{\label{sec:experiments}}

\begin{figure*} [ht]
    \centering
    \def\svgwidth{1\textwidth}
    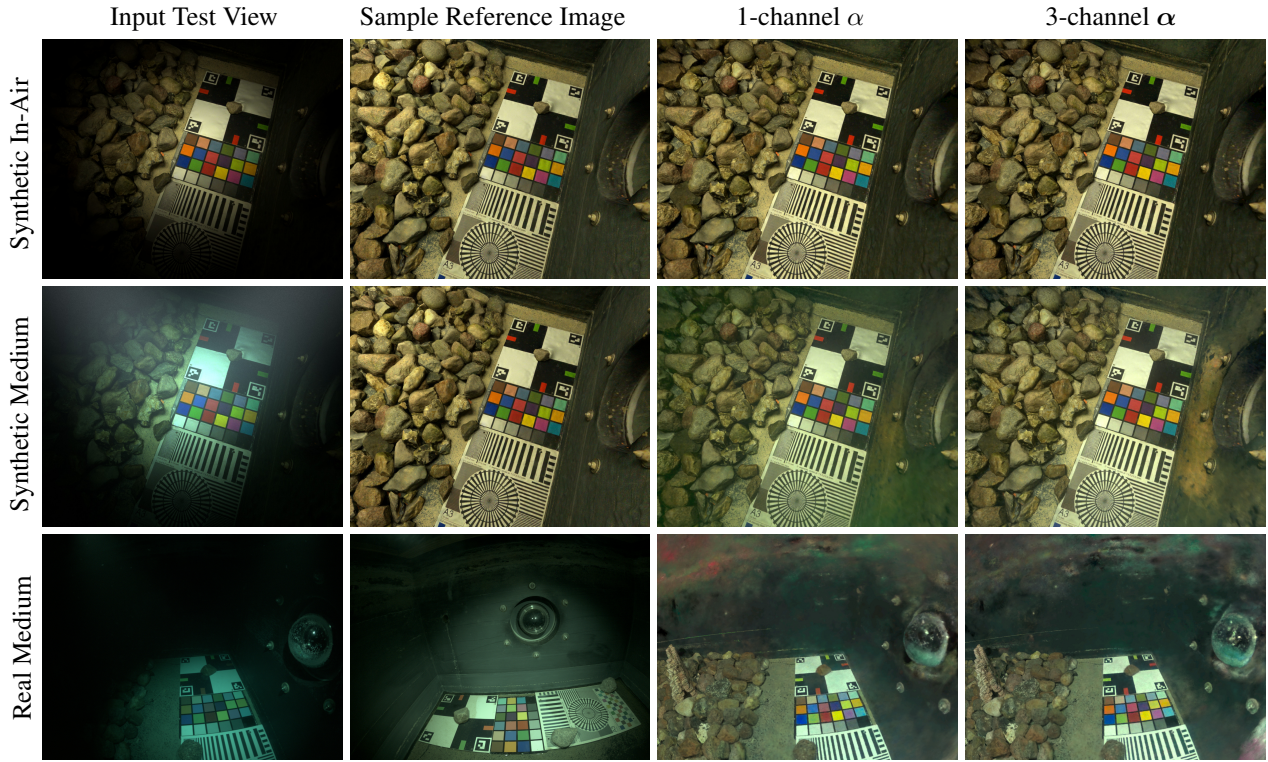
    \caption{Ablation of single- and three-channel illumination field. Estimating three channels for the illumination field has is advantageous for in medium data. It is mostly irrelevant for in-air data. To represent the received light at a certain scene point the field must be able to model the channel dependent attenuation of the light on its way to the object.}
    \label{fig:ablation_channels}
\end{figure*}

We evaluate our proposed method on synthetic and real datasets. Synthetic data is rendered using physically based ray-tracer Mitsuba3 \cite{nimier2019mitsuba}. Real data is captured in RAW using a GoPro Hero9 Black, enclosed in a dome port housing\footnote{The NeRF formulation of tracing back a camera ray into the scene easily allows consideration of refraction for a camera fixed inside a housing. This involves only one-time application of Snell's law and is much simpler that learning complex refraction patterns such as~\cite{zhan23nerfac}. Our approach therefore also supports refractive cameras~\cite{she24rsfm} at no extra cost, but the dome port housing used here avoids refraction in the first place.}~\cite{she22dome} with two co-moving light sources. 
To make the medium effects visible in a small $1m \times 2m$ tank, scattering agent and dye were added to the water to mimic the effects in clear sea water. For visualization RAW images are debayered and all images are sRGB gamma corrected. 

\subsection{Co-Moving Light and Medium Removal}

Results of applying the proposed method on different datasets are shown in Fig.~\ref{fig:eval_method}. The two first datasets do not contain any medium and the model was trained accordingly by disabling $\d C^{\mathrm{med}}(\d r(t))$ in Eq.~\ref{eq:final_model}. Note that DarkGS requires a human-in-the-loop light calibration procedure on a calibration board beforehand, while we simply train our model on their provided data. Additionally the compared to methods model only a single known light source, while we can model arbitrary light configurations. The datasets in Fig.~\ref{fig:eval_method} contain one, four, two and two lights from top to bottom. The method named Raw-Nerfacto is our integration of RawNeRF into the Nerfacto model. 
Due to a lack of competitor methods for our challenging scenario neither of the methods marked with an asterisk explicitly model co-moving light, and consequently the resulting scene contains heavy artifacts. SeaThru-NeRF does show some color restoration but can also not deal with the co-moving light. Parts of the scene which never receive light can obviously not be recovered. This can happen easily since the light cone effectively reduces the field of view of the camera.  The dark artifacts in the upper part of the clean \textit{Tank} view in \cref{fig:eval_method} or the ones visible in the bottom row of \cref{fig:ablation_channels} are caused by this.

\subsection{Ablation: Multi-channel Illumination Field}

\begin{table} [b]
  \centering
  \footnotesize
  \scalebox{1}{
  \begin{tabular}{@{}lccc@{}}
    \toprule
    & Synthetic In-Air& Synthetic Medium& Real Medium\\
    \midrule
    1-channel $\alpha$& 9.554& 12.780& 28.944\\
    3-channel $\d \alpha$& 10.143& 11.879& 30.077 \\
    \bottomrule
  \end{tabular}  
   }
  \caption{Ablation results of 1- versus 3-channel $\d \alpha$. Metric is the color difference as $L^2$ norm $\downarrow$ in pixels (0-255) between reference and input colors over all 24 color patches in linear color space.}
  \label{tab:ablation_channels}
\end{table}

\begin{figure*} [t]
    \centering
    \def\svgwidth{0.9\textwidth}
    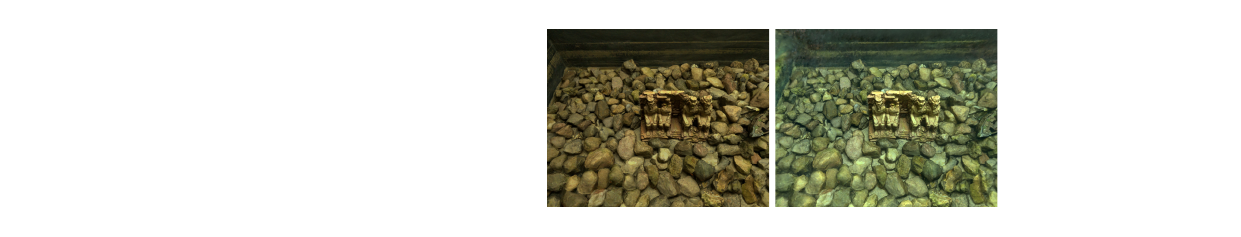
    \caption{Insufficient input data. Problematic camera trajectory at constant distance to the scene (a), ground truth image (b), reference clean view (c), intermixed medium effects in the object color (d), view dependency and illumination ambiguity (e). For this limited data, image (d) could only be separated from the illumination by training with disabled view dependency of the scene NeRF.}
    \label{fig:discussion}
\end{figure*}

Light color and spectral response of the camera are consistent over all views and can therefore not be disentangled, given just this image data. They will effectively be 'baked' into the object color. The relative illumination field can only learn the amount of light a point in space receives in relation to the camera frame. For in-air data this means that it is sufficient to estimate a single $\alpha = \mathcal{F}^l_\Theta(\phi_\mathrm{Hash}(\d x^c), \phi_\mathrm{SH}(\d n^c))$ representing this factor, assuming all co-moving light sources have the same color. In a medium the situation changes, because the light is already attenuated on its path to the object. This attenuation depends on the distance traveled from light source to object point and varies between wavelengths. We therefore estimate an $\alpha$ per color channel in this case. 

The relighting and color correction quality is quantified on the color checker chart present in the datasets. Color patches are extracted and averaged over five test views. The reference colors for the synthetic images are determined by rendering the same scene with constant, global illumination and no medium. For the real dataset we capture in-air images of the board using the same camera and light setup. To simulate a constant illumination with these exact lights, images are taken in a lawnmower-pattern while keeping a constant distance and angle. The highest intensity value of each color patch mean is then used as a reference for this patch. Because the object color can only be recovered up to scale all input and reference colors $\d c^i$, $\d c^r$ are aligned by a single factor estimated by minimizing $E$ in a least-squares fashion: $E = \sum_{j=0}^{24} |\d c^i_j - \d c^r_j|^2$ for all 24 colors of the board. The reported metric then is the mean $L^2$ norm between input and reference colors. Ablation renders are shown in Fig.~\ref{fig:ablation_channels}, metrics in Tab.~\ref{tab:ablation_channels}. While there is no visible difference for the in-air images, the single channel $\alpha$ images show a slight blue-green hue in medium. For the real data the metric does not show an improvement. We suspect that the rather small effect is drowned in the noise of the real dataset due to imperfect calibration and poses.

\subsection{Discussion}

While the proposed approach is very flexible in that it requires no calibration or knowledge of light sources and medium, it requires that this information is recoverable from the data. This involves mainly two aspects:
\paragraph{Observable Attenuation Changes}
The changes in color caused by the medium attenuation can only be observed if the distance to the scene varies. If camera poses are at a constant distance the degree of attenuation is also constant and can not be distinguished from the object color. Fig.~\ref{fig:discussion} shows an example of a problematic camera path where the distance to the scene does not vary much. As a result, the medium attenuation can not be estimated properly and the object color becomes mixed with medium effects. 
\paragraph{Observable Light Pattern}
Similarly, to discern object color from lighting effects the distribution of light in space must be constrained sufficiently. NeRF allows a certain degree of view dependent color variation typically meant to model specular highlights. This is usually implemented by providing the spherical harmonics encoded ray direction to the MLP and subsequently allows it to change the color based on viewing direction. With very limited poses this view dependency can interfere with the illumination field estimation. The result can be seen in the bottom right image in Fig.~\ref{fig:discussion} where the illumination effect has been absorbed into the scene MLP. For scenes with very uniform poses we find that disabling NeRF's ability to model view dependent variation can help to successfully estimate the illumination and separate it from the scene.

%% file: figs/eval/ablation_channels_figure.pdf_tex
\begingroup%
  \makeatletter%
  \providecommand\color[2][]{%
    \errmessage{(Inkscape) Color is used for the text in Inkscape, but the package 'color.sty' is not loaded}%
    \renewcommand\color[2][]{}%
  }%
  \providecommand\transparent[1]{%
    \errmessage{(Inkscape) Transparency is used (non-zero) for the text in Inkscape, but the package 'transparent.sty' is not loaded}%
    \renewcommand\transparent[1]{}%
  }%
  \providecommand\rotatebox[2]{#2}%
  \newcommand*\fsize{\dimexpr\f@size pt\relax}%
  \newcommand*\lineheight[1]{\fontsize{\fsize}{#1\fsize}\selectfont}%
  \ifx\svgwidth\undefined%
    \setlength{\unitlength}{496.85998535bp}%
    \ifx\svgscale\undefined%
      \relax%
    \else%
      \setlength{\unitlength}{\unitlength * \real{\svgscale}}%
    \fi%
  \else%
    \setlength{\unitlength}{\svgwidth}%
  \fi%
  \global\let\svgwidth\undefined%
  \global\let\svgscale\undefined%
  \makeatother%
  \begin{picture}(1,0.60379183)%
    \lineheight{1}%
    \setlength\tabcolsep{0pt}%
    \put(0.09513005,0.56993442){\color[rgb]{0,0,0}\makebox(0,0)[lt]{\lineheight{1.25}\smash{\begin{tabular}[t]{l}Input Test View\end{tabular}}}}%
    \put(0.28510579,0.56993442){\color[rgb]{0,0,0}\makebox(0,0)[lt]{\lineheight{1.25}\smash{\begin{tabular}[t]{l}Sample Reference Image\end{tabular}}}}%
    \put(0.80720444,0.56993442){\color[rgb]{0,0,0}\makebox(0,0)[lt]{\lineheight{1.25}\smash{\begin{tabular}[t]{l}3-channel $\d \alpha$\end{tabular}}}}%
    \put(0.03266093,0.39881628){\color[rgb]{0,0,0}\rotatebox{90}{\makebox(0,0)[lt]{\lineheight{0}\smash{\begin{tabular}[t]{l}Synthetic In-Air\end{tabular}}}}}%
    \put(0.57178586,0.56993442){\color[rgb]{0,0,0}\makebox(0,0)[lt]{\lineheight{1.25}\smash{\begin{tabular}[t]{l}1-channel $\alpha$\end{tabular}}}}%
    \put(0.03266092,0.19911494){\color[rgb]{0,0,0}\rotatebox{90}{\makebox(0,0)[lt]{\lineheight{0}\smash{\begin{tabular}[t]{l}Synthetic Medium\end{tabular}}}}}%
    \put(0.03461165,0.04047446){\color[rgb]{0,0,0}\rotatebox{90}{\makebox(0,0)[lt]{\lineheight{0}\smash{\begin{tabular}[t]{l}Real Medium\end{tabular}}}}}%
    \put(0,0){\includegraphics[width=\unitlength,page=1]{ablation_channels_figure.pdf}}%
  \end{picture}%
\endgroup%

%% file: figs/eval/discussion_v2.pdf_tex
\begingroup%
  \makeatletter%
  \providecommand\color[2][]{%
    \errmessage{(Inkscape) Color is used for the text in Inkscape, but the package 'color.sty' is not loaded}%
    \renewcommand\color[2][]{}%
  }%
  \providecommand\transparent[1]{%
    \errmessage{(Inkscape) Transparency is used (non-zero) for the text in Inkscape, but the package 'transparent.sty' is not loaded}%
    \renewcommand\transparent[1]{}%
  }%
  \providecommand\rotatebox[2]{#2}%
  \newcommand*\fsize{\dimexpr\f@size pt\relax}%
  \newcommand*\lineheight[1]{\fontsize{\fsize}{#1\fsize}\selectfont}%
  \ifx\svgwidth\undefined%
    \setlength{\unitlength}{595.27559055bp}%
    \ifx\svgscale\undefined%
      \relax%
    \else%
      \setlength{\unitlength}{\unitlength * \real{\svgscale}}%
    \fi%
  \else%
    \setlength{\unitlength}{\svgwidth}%
  \fi%
  \global\let\svgwidth\undefined%
  \global\let\svgscale\undefined%
  \makeatother%
  \begin{picture}(1,0.19047619)%
    \lineheight{1}%
    \setlength\tabcolsep{0pt}%
    \put(0,0){\includegraphics[width=\unitlength,page=1]{discussion_v2.pdf}}%
    \put(0.6290661,0.03036435){\color[rgb]{1,1,1}\makebox(0,0)[lt]{\lineheight{0}\smash{\begin{tabular}[t]{l}(d)\end{tabular}}}}%
    \put(0,0){\includegraphics[width=\unitlength,page=2]{discussion_v2.pdf}}%
    \put(0.01991315,0.03037255){\color[rgb]{0,0,0}\makebox(0,0)[lt]{\lineheight{0}\smash{\begin{tabular}[t]{l}(a)\end{tabular}}}}%
    \put(0.2609185,0.03036435){\color[rgb]{1,1,1}\makebox(0,0)[lt]{\lineheight{0}\smash{\begin{tabular}[t]{l}(b)\end{tabular}}}}%
    \put(0.444725,0.03037255){\color[rgb]{1,1,1}\makebox(0,0)[lt]{\lineheight{0}\smash{\begin{tabular}[t]{l}(c)\end{tabular}}}}%
    \put(0.81292608,0.03037255){\color[rgb]{1,1,1}\makebox(0,0)[lt]{\lineheight{0}\smash{\begin{tabular}[t]{l}(e)\end{tabular}}}}%
  \end{picture}%
\endgroup%

%% file: sec/5_conclusion.tex
\section{Conclusion}\label{sec:conclusion}

This work provides a novel extension to the NeRF framework enabling the recovery of a clean scene representation for data captured with co-moving lights. We also integrate a volumetric rendering formulation for a scattering medium, allowing the approach to work on images captured underwater. The model is very flexible and can deal with one or more light sources in arbitrary positions relative to the camera. The light distribution in space can be learned jointly with the medium effects making it the first approach capable of recovering a scene free from both effects without calibration. 
Currently shadows cast by the co-moving lights are not explicitly considered. Since the fully trained network contains information about light and scene geometry, extending the method to deal with dynamic shadows is a goal for future work.